# IMAGE ENHANCEMENT USING A GENERALIZATION OF HOMOGRAPHIC FUNCTION

VASILE PĂTRAȘCU [*]

***Abstract.*** *This paper presents a new method of gray level image enhancement, based on point transforms. In order to define the transform function, it was used a generalization of the homographic function.*

## I. Introduction

The image enhancement methods occupy an important place among the wide category of all image processing procedures. The great number of the existing methods is justified by the spread variety of images, which need specific methods [4]. In the specialty literature, one can find numerous methods [1,2,7]. This article describes a method of image enhancement, a member of point transform class [3]. The function used there for gray level transform is obtained by a generalization of the homographic function.

The structure of this paper is the following: section 2 includes a short description of the homographic function and its generalization using a new parameter; section 3 presents the fixing algorithm of interpolation points and, implicitly, the parameters of the generalized homographic function (used to transform the gray level in order to yield the enhanced image). The section 4 contains the experimental results, while in section 5 one can get some conclusions. Finally, the references are listed at the end of paper.

## II. The homographic function and its generalization

Let be the real homographic function:

$$f: R - \left\{-\frac{d}{c}\right\} \to R, c \neq 0, \quad f(x) = \frac{ax+b}{cx+d} \qquad (1)$$

[*] Department of Informatics Technology,
  Tarom Company,
  e-mail: vpatrascu@tarom.ro

The function parameters are calculated through 3 conditions; for example, knowing 3 interpolation points. Thus, if the values $f_1, f_2$ of the function $f$ in two points $x_1 < x_2$ are known, the function may be written as follows:

$$f(x) = \frac{\alpha_1 \cdot f_1 \cdot (x_2 - x) + \alpha_2 \cdot f_2 \cdot (x - x_1)}{\alpha_1 (x_2 - x) + \alpha_2 (x - x_1)}, \quad \alpha_1, \alpha_2 \in R \qquad (2)$$

Starting from (2) it was built *the generalized homographic function*, $g:[x_1, x_2] \rightarrow [g_1, g_2]$, introducing a new parameter $\gamma$. Thus:

$$g(x) = \frac{\alpha_1 \cdot g_1 \cdot (x_2 - x)^\gamma + \alpha_2 \cdot g_2 \cdot (x - x_1)^\gamma}{\alpha_1 \cdot (x_2 - x)^\gamma + \alpha_2 \cdot (x - x_1)^\gamma} \qquad (4)$$

where $g_1, g_2$ are the values of $g$ function in $x_1, x_2$. The $g$ function will be determined knowing the values $g_{c1}, g_{c2} \in [g_1, g_2]$ in two intermediary points $c_1, c_2 \in (x_1, x_2)$, namely $g(c_1) = g_{c1}$, $g(c_2) = g_{c2}$. With the last two conditions and, also with (4) it got two equations:

$$\begin{cases} \alpha_1 \cdot (g_1 - g_{c1}))(x_2 - c_1)^\gamma + \alpha_2 \cdot (g_2 - g_{c1})(c_1 - x_1)^\gamma = 0 \\ \alpha_1 \cdot (g_1 - g_{c2})(x_2 - c_2)^\gamma + \alpha_2 \cdot (g_2 - g_{c2})(c_2 - x_1)^\gamma = 0 \end{cases} \qquad (5)$$

From (5) we obtain for $\gamma$ the value:
$$\gamma_0 = \frac{\ln\left(\dfrac{g_{c1} - g_1}{g_2 - g_{c1}} \cdot \dfrac{g_2 - g_{c2}}{g_{c2} - g_1}\right)}{\ln\left(\dfrac{c_1 - x_1}{x_2 - c_1} \cdot \dfrac{x_2 - c_2}{c_2 - x_1}\right)} \qquad (6)$$

In the end, function $g$ has the next relation as it results from (5) using (4):

$$g(x) = \frac{\dfrac{g_2 - g_{c1}}{(x_2 - c_1)^{\gamma_0}} \cdot g_1 \cdot (x_2 - x)^{\gamma_0} + \dfrac{g_{c1} - g_1}{(c_1 - x_1)^{\gamma_0}} \cdot g_2 \cdot (x - x_1)^{\gamma_0}}{\dfrac{g_2 - g_{c1}}{(x_2 - c_1)^{\gamma_0}} \cdot (x_2 - x)^{\gamma_0} + \dfrac{g_{c1} - g_1}{(c_1 - x_1)^{\gamma_0}} \cdot (x - x_1)^{\gamma_0}} \qquad (7)$$

where $\gamma_0$ is the solution (6).

### III. The interpolation points algorithm

We will consider as representation for the gray level space the set $E = [0,1]$. A gray level image is described by its function of gray level $l: D \rightarrow E$ where $D \subset R^2$ is a compact set and represent the image support. The interpolation points $x_1, c_1, c_2, x_2$ will be *the interwoven means* of the image $l$ [6], and the values $g_1, g_{c1}, g_{c2}, g_2$ will be *the interwoven means* of an image with an uniform distribution in the gray levels set $E$. We will particularize the shown method in [6] for the discrete case with four values. Thus:

$$x_1 = \min_{(x,y) \in D} l(x,y), \quad x_2 = \max_{(x,y) \in D} l(x,y) \tag{8}$$

The points $c_1, c_2$ will be determinate from the following two equations system:

$$c_i = \frac{1}{card(D_i)} \sum_{(x,y) \in D_i} l(x,y) \quad \text{for } i = 1,2 \tag{9}$$

where $D_1 = \{(x,y) \in D \mid l(x,y) \in [x_1, c_2]\}, D_2 = \{(x,y) \in D \mid l(x,y) \in [c_1, x_2]\}$ (10) and $card(D_i)$ is the cardinality of $D_i$. As it is difficult to find an analytical solution for (9), the problem is numerically solved. Suppose that the images' gray levels have values in the interval $[l_{min}, l_{max}]$. The next procedure will be used for interpolation points calculus [6]:

1. Initialisation: choose $\varepsilon$ the constant for stopping the procedure, $m = 0$, $x_1 = l_{min}$, $x_2 = l_{max}$, and also the initial values:

$$c_1^{(0)} = \frac{l_{min} + l_{med}}{2}, \quad c_2^{(0)} = \frac{l_{med} + l_{max}}{2} \quad \text{where} \quad l_{med} = \frac{1}{card(D)} \sum_{(x,y) \in D} l(x,y)$$

2. One calculate $D_1^{(m)} = \{(x,y) \in D \mid l(x,y) \in [x_1, c_2^{(m)}]\}$,
$D_2^{(m)} = \{(x,y) \in D \mid l(x,y) \in [c_1^{(m)}, x_2]\}$

and 
$$c_i^{(m+1)} = \frac{1}{card(D_i^{(m)})} \sum_{(x,y) \in D_i^{(m)}} l(x,y) \quad \text{for } i = 1,2.$$

3. If $\left| c_i^{(m+1)} - c_i^{(m)} \right| < \varepsilon$ for $i = 1,2$ then pass to the step 4, else $m = m + 1$ and go to the step 2.

4. Save the results $x_1, c_1 = c_1^{(m+1)}, c_2 = c_2^{(m+1)}, x_2$ and stop.

The described algorithm gives equidistant points when the image has an uniform distribution for the gray levels.

## IV. Experimental results

The proposed method was used to enhance some images. The interpolation nodes are chosen so that the new image has a gray level distribution close to an uniform one. Thus, if $[0,1]$ is the interval of gray levels then the values $g_1, g_{c1}, g_{c2}, g_2$ of the interpolation function are equidistant and yield from:

$$g_1 = 0, \, g_{c1} = \frac{1}{3}, \, g_{c2} = \frac{2}{3}, \, g_2 = 1 \tag{11}$$

To exemplify, two images were picked out: "miss" in Fig.1a and "lax" in Fig.2a. Their histograms are in Fig.1b and Fig.2b. The image "miss" has the following interpolation parameters: $\gamma = 1.7, \alpha_1 = 2.4, \alpha_2 = 3.6$. Analogous for "lax" image the following were obtained: $\gamma = 5, \alpha_1 = 25.6, \alpha_2 = 86.9$. Their graphics are shown in Fig.1c for "miss", and Fig.2c for "lax". The enhanced images can be seen in Fig.1d and Fig.2d, also their histogram in Fig.1e and Fig.2e.

## V. Conclusions

The paper presented a method for enhancing the gray level images. The method is based on point transforms defined by interpolation functions that are realised with the generalized homographic function. These functions are determined by simple formulae, which need short calculus time. In establishing the interpolation points it was chosen an algorithm that considers the statistics properties of gray level images [6]. Their points' computing has a good convergence and requires few iterations. Future perspectives for the shown method could be: 1) the extension for color images; 2) using the *means of k-order* for computing the points of interpolation [5].

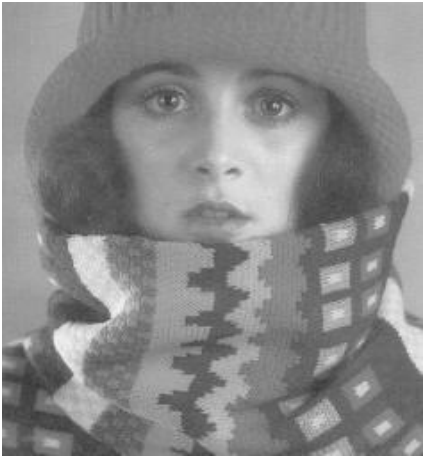
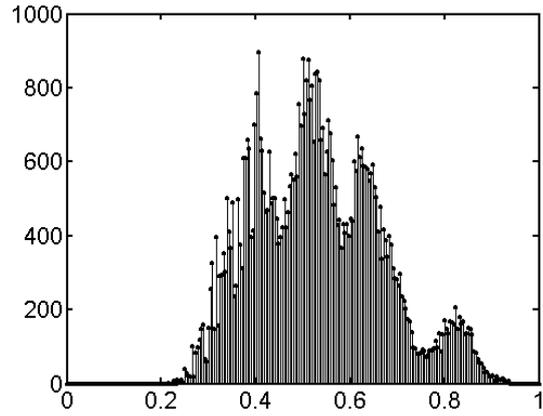

a)                                                  b)

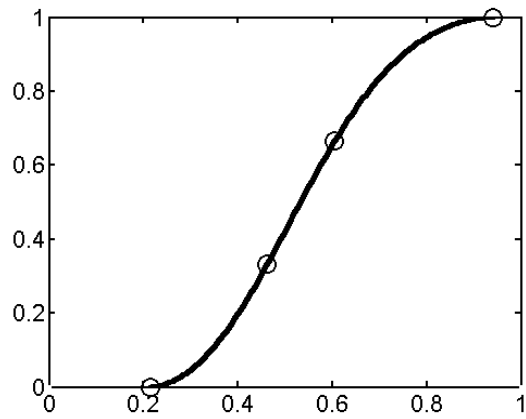

c)

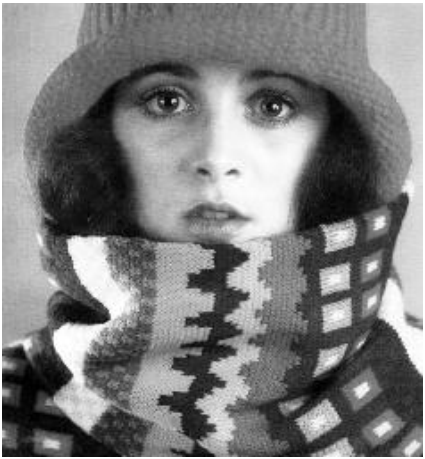
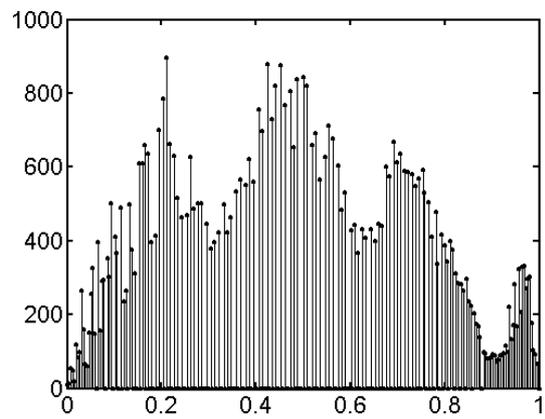

d)                                                  e)

Fig.1. a) Original image "miss", b) The gray level histogram of original image, c) The gray level transform (*the generalized homographic function*) where the circles represent the interpolation nodes, d) The enhanced image, e) The gray level histogram of enhanced image.

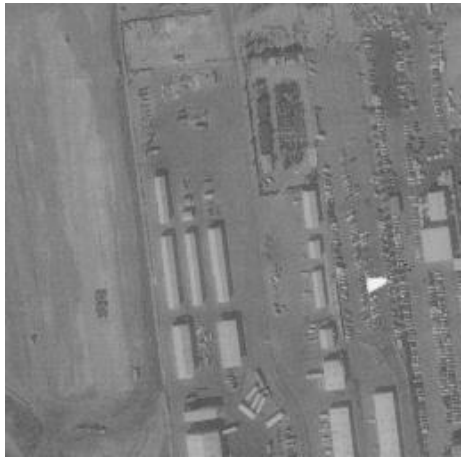
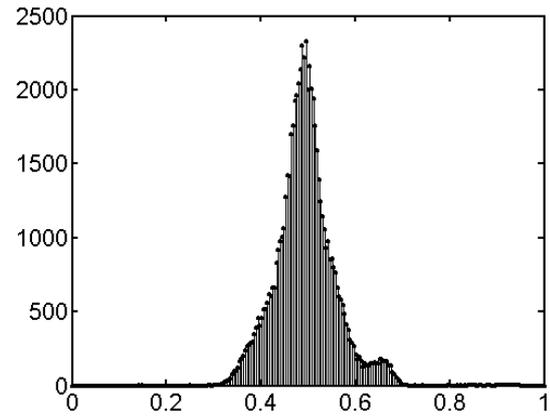

a) b)

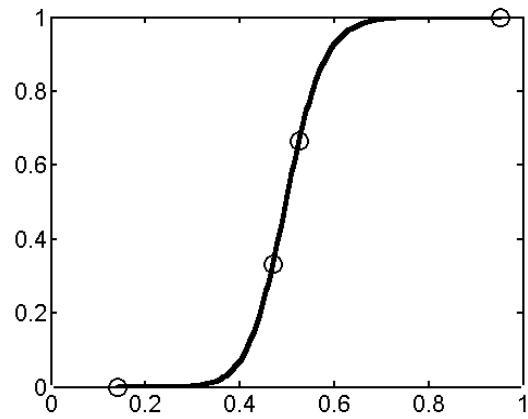

c)

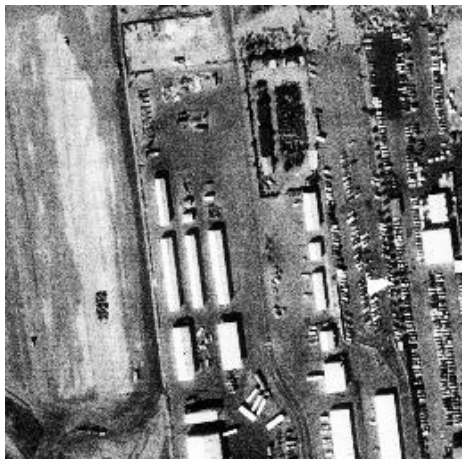
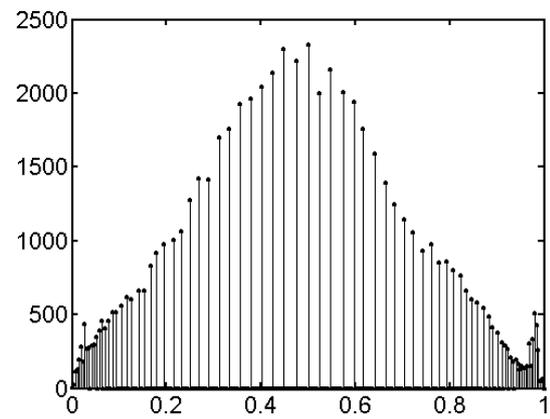

d) e)

Fig.2. a) Original image "lax", b) The gray level histogram of original image, c) The gray level transform (*the generalized homographic function*) where the circles represent the interpolation nodes, d) The enhanced image, e) The gray level histogram of enhanced image.